\documentclass[review]{elsarticle}

\usepackage{fullpage}
\usepackage{amsfonts,amssymb,amsmath}
\usepackage{graphicx}
\usepackage{subfig}

\usepackage{lineno}
\usepackage{textgreek}
\usepackage{fancyvrb}
\def\bo{\mbox{\raise .35em \hbox{\underline{\scriptsize o}\ }}}
\def\ba{\mbox{\raise .35em \hbox{\underline{\scriptsize a}\ }}}
\def\Exmo{Ex\mbox{\raise .35em \hbox{\underline{\footnotesize mo}\ }}}

\def\Exmos{Ex\mbox{\raise .35em \hbox{\underline{\footnotesize mos}\ }}}
\def\Excia{Ex\mbox{\raise .35em \hbox{\underline{\footnotesize cia}\ }}}

\def\trace{\mathop{\mathrm{trace}}}

\usepackage{changes}

\usepackage{lineno}

\newcommand*\patchAmsMathEnvironmentForLineno[1]{%
  \expandafter\let\csname old#1\expandafter\endcsname\csname #1\endcsname
  \expandafter\let\csname oldend#1\expandafter\endcsname\csname end#1\endcsname
  \renewenvironment{#1}%
  {\linenomath\csname old#1\endcsname}%
  {\csname oldend#1\endcsname\endlinenomath}}%
  \newcommand*\patchBothAmsMathEnvironmentsForLineno[1]{%
    \patchAmsMathEnvironmentForLineno{#1}%
    \patchAmsMathEnvironmentForLineno{#1*}}%
    \AtBeginDocument{%
      \patchBothAmsMathEnvironmentsForLineno{equation}%
      \patchBothAmsMathEnvironmentsForLineno{align}%
      \patchBothAmsMathEnvironmentsForLineno{flalign}%
      \patchBothAmsMathEnvironmentsForLineno{alignat}%
      \patchBothAmsMathEnvironmentsForLineno{gather}%
      \patchBothAmsMathEnvironmentsForLineno{multline}%
    }

\begin{document}

    \begin{frontmatter}

      \title{Pl\"{u}cker Correction Problem: Analysis and Improvements in Efficiency}

      \author[rvt,focal]{Jo\~{a}o~R.~Cardoso}
      \ead{jocar@isec.pt}
      \author[els]{Pedro~Miraldo\fnref{fn3}\corref{cor1}}
      \ead{pmiraldo@isr.tecnico.ulisboa.pt}
      \author[rvt]{Helder~Araujo}
      \ead{helder@isr.uc.pt}

      \cortext[cor1]{Corresponding author}
  % \cortext[cor2]{Principal corresponding author}
  % \fntext[fn1]{This is the specimen author footnote.}
  % \fntext[fn2]{Another author footnote, but a little more longer.}
      \fntext[fn3]{P. Miraldo was supported by a Post-Doctoral Grant from the EC Project RoCKIn (FP7-ICT-601012).}

      \address[rvt]{Institute for Systems and Robotics, Universidade de Coimbra, Coimbra, Portugal}
      \address[focal]{Coimbra Institute of Engineering, Instituto Polit\'{e}cnico de Coimbra, Coimbra, Portugal}
      \address[els]{Institute for Systems and Robotics (LARSyS), Instituto Superior T\'{e}cnico, Universidade de Lisboa, Portugal}

      \begin{abstract}
        A given six dimensional vector represents a 3D straight line in Pl\"{u}cker coordinates if its coordinates satisfy the Klein quadric constraint. In many problems aiming to find the Pl\"{u}cker coordinates of lines, noise in the data and other type of errors contribute for obtaining 6D vectors that do not correspond to lines, because of that constraint. A common procedure to overcome this drawback is to find the Pl\"{u}cker coordinates of the lines that are closest to those vectors. This is known as the Pl\"{u}cker correction problem. In this article we propose a simple, closed-form, and global solution for this problem. When compared with the state-of-the-art method, one can conclude that our algorithm is easier and requires much less operations than previous techniques (it does not require Singular Value Decomposition techniques).
      \end{abstract}


    \end{frontmatter}

%    \linenumbers

    \section{Introduction}

%% intro
    In many 3D vision problems that range from camera calibration to robot navigation, it is required to represent 3D straight lines. Pl\"{u}cker coordinates are one of the most used formulations \cite{article:pottman:2001}. Vectors of Pl\"{u}cker coordinates are built-up by stacking both direction and moment of the respective lines (both 3D vectors), giving a 6D vector. Direction and moment vectors must satisfy the so-called Klein quadric constraint, that is, they need to be orthogonal to each other. In this paper we address the problem of estimating P\"{u}cker coordinates from general (unconstrained) $\mathbb{R}^6$ vectors. This is called the Pl\"{u}cker correction problem.

    For many reasons, especially when considering data with noise, it is hard to include the orthogonal constraint in the estimator (frequently, it requires non-linear procedures). Some authors do not consider this constraint on their methods or propose optimization techniques involving this constraint but, to avoid unnecessary computational effort, non-linear procedures are stopped before fulfilling the respective constraint. Several examples can be found in the literature, e.g. camera models (mapping between pixels and 3D straight lines) \cite{article:miraldo:2013}; triangulation of 3D lines \cite{article:josephson:2008}; structure-from-motion using lines \cite{Bartoli05,article:lemaire:2007}; and 3D reconstruction of lines using a single image of a non-central catadioptric camera \cite{article:lemaire:2007}. Thus, to get Pl\"{u}cker coordinates on these cases, a Pl\"{u}cker correction needs to be applied.

%% state-of-the-art solutions
    The state-of-art method for solving the Pl\"{u}cker correction problem is due to Bartoli and Sturm~at~\cite{Bartoli05}, hereafter called BS method. They find the closest Pl\"{u}cker coordinates (in Euclidean sense) from an unconstrained six-dimensional vector by orthogonally projecting the input vector onto the Klein quadric (and so it verifies the orthogonal constraint). Their approach involves SVD decompositions and can be found in \cite[p. 425]{Bartoli05}. It should be noticed that in the description of the BS Pl\"{u}cker correction algorithm (Table 2 in \cite{Bartoli05}) there is a typo in the entries of the matrix $T$. The correct matrix $T$ is given by
    \begin{equation}
      T=\left[\begin{array}{rr}
          z_{12} & z_{22}\\
          z_{21} & -z_{11}
      \end{array}\right].
    \end{equation}
    We performed a detailed analysis of the proof of the BS method (\cite[p. 425]{Bartoli05}) and propose some clarifications. For the proof of BS method the following identity is used:
    \begin{equation}\label{identity}
      \|{\mathbf U}{\mathbf A}-{\mathbf B}\|=\|{\mathbf A}-{\mathbf U}^T{\mathbf B}\|
    \end{equation}
    where  ${\mathbf A}$ and ${\mathbf B}$ are matrices of sizes $n\times n$ and $m\times n$, respectively, with  $m>n$, and ${\mathbf U}$ is an $m\times n$ matrix with orthonormal columns (i.e . ${\mathbf U}^T{\mathbf U}={\mathbf I}$, but ${\mathbf U}{\mathbf U}^T\neq {\mathbf I}$). This identity is related with the invariance of the Frobenius norm for left multiplication by matrices with orthonormal columns.

    However, (\ref{identity}) does not hold in general. It is valid if, and only if, the space of columns of ${\mathbf B}$ coincides with the space of columns of ${\mathbf U}$, in which case ${\mathbf B}={\mathbf U}{\mathbf F}$, for some matrix ${\mathbf F}$ of size $n\times n$. We notice that, in the proof of the BS method, this requirement is not explicitly mentioned.

%Moreover, in the proof of BS method (see \cite[p. 425]{Bartoli05}) there are some claims that are not entirely correct, which are related %with the invariance of the Frobenius norm for left multiplication by matrices with orthonormal columns. Assuming that $m>n$, let ${\mathbf %A}$ and ${\mathbf B}$ be matrices of sizes $n\times n$ and $m\times n$, respectively. Does the identity
%\begin{equation}\label{identity}
%\|{\mathbf U}{\mathbf A}-{\mathbf B}\|=\|{\mathbf A}-{\mathbf U}^T{\mathbf B}\|
%\end{equation}
%hold for any $m\times n$ matrix ${\mathbf U}$ with orthonormal columns (i.e. ${\mathbf U}^T{\mathbf U}={\mathbf I}$, but ${\mathbf %U}{\mathbf U}^T\neq {\mathbf I}$)? It turns out that this is not true, in general. However, if the space of columns of ${\mathbf B}$ %coincides with the space of columns of ${\mathbf U}$, in which case ${\mathbf B}={\mathbf U}{\mathbf F}$, for some matrix ${\mathbf F}$ of %size $n\times n$, then (\ref{identity}) holds. We notice that in the proof of the BS method we are led to wrongly conclude that %(\ref{identity}) is true for any orthonormal matrix ${\mathbf U}$.

    Paper \cite{Wu15} also addresses the same problem. However they do not address the problem
    of the computational efficiency and the proposed approach is quite different from the
    method described in this paper. Moreover, even if the solution proposed in
    \cite{Wu15} is indeed a global minimum, there is no formal proof of such a fact. We shall recall that Lagrange multipliers yield only local minima, unless  the existence of a global minimum is guaranteed.

%One of the referees of this paper pointed out the paper \cite{Wu15} that also uses Lagrange multipliers techniques for the Pl\"{u}cker %correction problem. Such a paper has been submitted at the same time of ours but the approach included there is quite different. The %formulae proposed in \cite{Wu15} is indeed a global minimum, but this issue is not proved there. We recall that Lagrange multipliers give %only local minima.

%%% Our approach.... (also outline of the paper)
%In this paper we propose a simple and explicit formula for the Pl\"{u}cker correction problem, Sec.~\ref{explicit}. To derive such a formula we use Lagrange multipliers techniques: the associated first order optimality conditions are solved and the local minimum that gives the smallest value in the objective function is identified, Sec.~\ref{local}. In Sec.~\ref{global} we prove theoretically that our formula gives a global minimum. A comparison between the proposed method and the state-of-the-art method of \cite{Bartoli05} is carried out in Sec.~\ref{experiments} and some conclusions are drawn in Sec.~\ref{sec:conclusions}.

    \subsection{Notations and Problem Definition}

    Column vectors are represented by bold small letters ({\it e.g.~} $\mathbf{a}\in\mathbb{R}^n$ for an $n$-dimensional vector). Bold capital letters denote matrices ({\it e.g.~} $\mathbf{A}\in\mathbb{R}^{n\times m}$ for an $n\times m$ matrix). Regular small letters denoted zero dimensional elements ({\it e.g.~} $a$). $\|.\|$ denotes the Frobenius norm for matrices or the $2$-norm for vectors. Recall that, for any matrix $\mathbf{A}$, the Frobenius norm is given by $\| \mathbf{A} \|^2=\trace(\mathbf{A}^T\mathbf{A})$ and, for any vector ${\mathbf u}$, the $2$-norm satisfies $\|{\mathbf u}\|^2={\mathbf u}^T{\mathbf u}$.

    The Pl\"{u}cker coordinates of a 3D straight line ${\cal G}$ can be represented by a six dimensional vector:
    \begin{equation}
      {\cal G} \sim ({\mathbf u}^T,\mathbf{v}^T) \in \mathbb{R}^{6},
    \end{equation}
    where ${\mathbf u}, \mathbf{v}\in\mathbb{R}^{3}$ are, respectively, the direction and moment of the line, verifying the Klein quadric constraint
    \begin{equation}{\mathbf u}^T \mathbf{v}={\mathbf 0},\end{equation} see \cite{article:pottman:2001}.

%Due to this constraint, in general, a given six dimensional vector does not represent a 3D line. For this reason, in some problems affected with noise, the computed approximation $\overline{\cal G}$ for the 3D line may not correspond to a line. A procedure that is commonly used to overcome this drawback is the so-called Pl\"{u}cker correction, which aims to find the closest 3D line to $\overline{\cal G}$.

    Let ${\mathbf a}$ and ${\mathbf b}$ be given vectors in $\mathbb{R}^{3}$ (not necessarily satisfying the orthogonality constraint), and assume that ${\mathbf x}$ and ${\mathbf y}$ denote vectors in $\mathbb{R}^{3}$. Mathematically, the Pl\"{u}cker correction problem corresponds to solving the nonlinear constrained optimization problem formulated as
    \begin{equation}\label{min-prob}
      \min_{{\mathbf x}^T{\mathbf y}={\mathbf 0}} \left\|[{\mathbf a}\ {\mathbf b}]-[{\mathbf x}\ {\mathbf y}]\right\|^2.
    \end{equation}
   % As mentioned before, an algorithm for solving (\ref{min-prob}) is proposed in \cite[Sec. 4.2]{Bartoli05}, but in many cases, the solution produced is not actually ``the closest'' one (i.e., the global minimum). However, it satisfies the Klein quadric constraint, which means that it corresponds to a 3D line.
   While the objective function
    \begin{equation}\label{obj-function}
      f({\mathbf x},{\mathbf y})\doteq\|{\mathbf a}-{\mathbf x}\|^2+\|{\mathbf b}-{\mathbf y}\|^2,
    \end{equation}
    is convex, the Klein quadric constraint ${\mathbf x}^T{\mathbf y}={\mathbf 0}$ is not. The optimization problem (\ref{min-prob}) belongs to a class of non-convex problems known in the literature as quadratically constrained quadratic programs. This means, in particular, that the existence of a global minimum may be a non trivial problem.

    The main goals of this paper are to prove the existence of a global minimum for (\ref{min-prob}) and to give an explicit formula for computing such a minimum. Our approach is essentially based on the application of the classical Lagrange multipliers to the constrained problem (\ref{min-prob}). It does not involve singular value decompositions which turns it faster than the BS method. Our results are supported by mathematical proofs and numerical experiments. In addition our method is designed to deal with general $n$-dimensional vectors ${\mathbf a}$ and ${\mathbf b}$, and not exclusively with $3$-dimensional vectors.

    \subsection{The Explicit Formula}
    \label{explicit}

    Let us consider two general 3D vectors $\mathbf{a}$ and $\mathbf{b}$, such that ${\mathbf a} \neq \pm {\mathbf b}$ and ${\mathbf b}\neq {\mathbf 0}$. As it will be shown in the following sections, the proposed solution $({\mathbf x}_\ast,\,{\mathbf y}_\ast)$ for the global minimum of (\ref{min-prob}) is given by
    \begin{equation}
      \label{eq:explicit:solution}
      {\mathbf x}_\ast=\frac{1}{1-\alpha^2}\left({\mathbf a}-\alpha{\mathbf b}\right) \;\; \text{and} \;\;\; {\mathbf y}_\ast=\frac{1}{1-\alpha^2}\left({\mathbf b}-\alpha{\mathbf a}\right),\\
    \end{equation}
    where
    \begin{equation}
      \alpha \doteq \frac{2p}{q+\sqrt{q^2-4p^2}},\ \text{with}\ p \doteq {\mathbf a}^T{\mathbf b},\ \text{and}\ q\doteq\|{\mathbf a}\|^2+\|{\mathbf b}\|^2.
    \end{equation}

    \section{Pl\"{u}cker Correction using Lagrange Multipliers}\label{local}
    Suppose again that ${\mathbf a}$ and ${\mathbf b}$ satisfy ${\mathbf a}\neq \pm {\mathbf b}$ and ${\mathbf b}\neq {\mathbf 0}$, and let ${\cal L}({\mathbf x},{\mathbf y},\lambda)$ denote the Lagrangian function associated to (\ref{min-prob}). Some calculation yields
    \begin{equation}\label{lagrangian}
      {\cal L}({\mathbf x},{\mathbf y},\lambda)=\|{\mathbf a}\|^2+\|{\mathbf b}\|^2-2\,{\mathbf a}^T{\mathbf x}-2\,{\mathbf b}^T{\mathbf y}+\|{\mathbf x}\|^2+
      \|{\mathbf y}\|^2+2\,\lambda\, {\mathbf x}^T{\mathbf y},
    \end{equation}
    where the real number $\lambda$ is the Lagrange multiplier, $\mathbf{x}$ and $\mathbf{y}$ are the aimed solutions. The partial derivatives of the Lagrangian with respect to ${\mathbf x}$, ${\mathbf y}$ and $\lambda$ are (for formulae of derivatives with respect to vectors see \cite{Lutkepohl96})
    \begin{align}
      \frac{\partial {\cal L}}{\partial {\mathbf x}} =&~-2{\mathbf a}+2{\mathbf x}+2\lambda{\mathbf y}\\
      \frac{\partial {\cal L}}{\partial {\mathbf y}} =&~-2{\mathbf b}+2{\mathbf y}+2\lambda{\mathbf x}\\
      \frac{\partial {\cal L}}{\partial \lambda} =&~2{\mathbf x}^T{\mathbf y}.
    \end{align}
    Equating these partial derivatives to zero, one obtains the first order optimality conditions (also known as Karush, Kuhn, Tucker conditions; see \cite{Luenberger08,Nocedal99}):
    \begin{eqnarray}
      {\mathbf a}-{\mathbf x}&=&\lambda {\mathbf y}\label{kkt-eq1}\\
      {\mathbf b}-{\mathbf y}&=&\lambda {\mathbf x}\label{kkt-eq2}\\
      {\mathbf x}^T{\mathbf y}&=&0.\label{kkt-eq3}
    \end{eqnarray}
    From (\ref{kkt-eq1}) and (\ref{kkt-eq2}),
    \begin{equation}\label{x-y}
      {\mathbf x}=\frac{{\mathbf a}-\lambda{\mathbf b}}{1-\lambda^2},\quad \mbox{and}\quad {\mathbf y}=\frac{{\mathbf b}-\lambda{\mathbf a}}{1-\lambda^2}
    \end{equation}
    (a geometric interpretation of (\ref{kkt-eq3}) and (\ref{x-y}) can be found in Figure \ref{figure1}).
    Replacing ${\mathbf x}$ and ${\mathbf y}$ in (\ref{kkt-eq3}), leads to the quadratic equation in $\lambda$
    \begin{equation}\label{quadratic}
      p\lambda^2-q\lambda+p=0,
    \end{equation}
    where $p$ and $q$ are
    \begin{equation}
      p \doteq{\mathbf a}^T{\mathbf b} \;\; \text{and} \;\; q \doteq\|{\mathbf a}\|^2+\|{\mathbf b}\|^2.
    % & \alpha&\doteq&\frac{2p}{q+\sqrt{q^2-4p^2}}.\nonumber\\
    \end{equation}
    Hence, the solutions of (\ref{quadratic}) are
    \begin{equation}\label{lambda}
      \lambda_1\doteq\frac{q+\sqrt{q^2-4p^2}}{2p}\quad \mbox{and}\quad  \lambda_2\doteq\frac{q-\sqrt{q^2-4p^2}}{2p}.
    \end{equation}
    Denoting, for $i=1,2$,
    \begin{equation}\label{xi-yi}
      \left({\mathbf x}_i,{\mathbf y}_i\right)=\left(\frac{{\mathbf a}-\lambda_i{\mathbf b}}{1-\lambda_i^2},\, \frac{{\mathbf b}-\lambda_i{\mathbf a}}{1-\lambda_i^2}\right),
    \end{equation}
    we know that $\left({\mathbf x}_1,{\mathbf y}_1\right)$ and $\left({\mathbf x}_2,{\mathbf y}_2\right)$ are the candidates to be a local minimum of the Lagrangian ${\cal L}({\mathbf x},{\mathbf y},\lambda)$. However, since the gradient of the constraint ${\mathbf x}^T{\mathbf y}={\mathbf 0}$ involved in (\ref{min-prob}) annihilates at $\left({\mathbf 0},{\mathbf 0}\right)$, one also needs to consider this non regular and non stationary point. Thus a local minimum of (\ref{min-prob}) must be attained at one of the following three points: $\left({\mathbf x}_1,{\mathbf y}_1\right)$, $\left({\mathbf x}_2,{\mathbf y}_2\right)$ or $\left({\mathbf 0},{\mathbf 0}\right)$.

    Now we shall note the following facts.
    \begin{itemize}
      \item The assumption of ${\mathbf a}$ and ${\mathbf b}$ being such that ${\mathbf a}\neq \pm {\mathbf b}$, with ${\mathbf b}\neq {\mathbf 0}$, guarantees, in particular, that $\lambda$ cannot be $\pm 1$, which ensures that ${\mathbf x}$ and ${\mathbf y}$ in (\ref{x-y}) are well defined.
      \item In (\ref{lambda}), $\lambda_i$ is always a real number because $q\geq 2p$. Indeed,
        $$q-2p={\mathbf a}^T{\mathbf a}+{\mathbf b}^T{\mathbf b}-2{\mathbf a}^T{\mathbf b}=({\mathbf a}-{\mathbf b})^T({\mathbf a}-{\mathbf b})=\|{\mathbf a}-{\mathbf b}\|^2\geq 0.$$
        Moreover, if $p=0$, then ${\mathbf a}$ and ${\mathbf b}$ are orthogonal, which means that the global minimum of the objective function is attained at $\left({\mathbf a},{\mathbf b}\right)$.
    \end{itemize}

    We end this section by showing that the objective function (\ref{obj-function}) satisfies
    \begin{equation}\label{ineq}
      f\left({\mathbf x}_2,{\mathbf y}_2\right)\leq f\left({\mathbf x}_1,{\mathbf y}_1\right)\leq f\left({\mathbf 0},{\mathbf 0}\right),
    \end{equation}
    which proves that $\left({\mathbf x}_2,{\mathbf y}_2\right)$ given in (\ref{xi-yi}) is the local minimum where the objective function $f$ attains the smallest value. In Sec.~\ref{global} is is shown that $\left({\mathbf x}_2,{\mathbf y}_2\right)$ is also the global minimum.
    Substituting ${\mathbf x}$ and ${\mathbf y}$ given in (\ref{x-y}) in the objective function (\ref{obj-function}), some calculation yields the following real function depending on the real variable $\lambda$:
    \begin{equation}\label{def-g}
      g(\lambda)\doteq \left(\frac{\lambda}{1-\lambda^2}\right)^2\left(q\lambda^2-4p\lambda +q\right).
    \end{equation}
    Now (\ref{ineq}) can be rewritten as
    \begin{equation}\label{ineq2}
      g(\lambda_2)\leq g(\lambda_1)\leq q.
    \end{equation}

    Let us write the function $g$ in the form
    \begin{equation}\label{g-phi}
      g(\lambda)=\left[\phi(\lambda)\right]^2\, \psi(\lambda),
    \end{equation}
    with $\phi(\lambda)\doteq \lambda/(1-\lambda^2)$ and $\psi(\lambda)\doteq q\lambda^2-4p\lambda +q$.

    For $\lambda_1$ and $\lambda_2$ given in (\ref{lambda}), it is not hard to check that $\phi(\lambda_1)=-\phi(\lambda_2)$, which implies  $\left[\phi(\lambda_1)\right]^2=\left[\phi(\lambda_2)\right]^2$. Hence, to prove the first inequality of (\ref{ineq2}), it remains to show that
    $\psi(\lambda_2)\leq \psi(\lambda_1)$.

    In fact, because $\lambda_i$ ($i=1,2$) satisfies the quadratic equation (\ref{quadratic}), one has $\lambda_i^2=(q/p)\lambda_i-1$. Therefore,
    $$\psi(\lambda_i)=\frac{q^2-4p^2}{p}\lambda_i,$$
    and, consequently,
    $$\psi(\lambda_2)=\frac{q(q^2-4p^2)-(q^2-4p^2)^{3/2}}{2p^2}\leq \psi(\lambda_1)=\frac{q(q^2-4p^2)+(q^2-4p^2)^{3/2}}{2p^2}.$$
    Finally, considering the equality $\lambda_1^2=(q/p)\lambda_1-1$ and the fact that the quadratic function $\psi(\lambda)$ is non negative, for all real numbers $\lambda$, the second inequality of (\ref{ineq2}) arises as a consequence of the equivalences
    \begin{eqnarray*}
      g(\lambda_1)\leq q & \Leftrightarrow & \lambda_1(-4p\lambda_1^2+3q\lambda_1)\leq q\\
      & \Leftrightarrow &  \lambda_1(4p-q\lambda_1)\leq q \\
      & \Leftrightarrow &  -\psi(\lambda_1)\leq 0.
    \end{eqnarray*}

    \subsection{Existence of a Global Minimum}\label{global}

    Given a unit vector ${\mathbf s}$ in the Euclidean space $\mathbb{R}^n$, consider the line $\ell$ through the origin with the direction of ${\mathbf s}$. Given another vector ${\mathbf a}\in\mathbb{R}^n$, it is well-known that the unique vector of $\ell$ that is closest to ${\mathbf a}$ (in Euclidean sense) is the orthogonal projection of ${\mathbf a}$ onto $\ell$, which is the vector ${\mathbf x}\doteq({\mathbf a}^T{\mathbf s}){\mathbf s}$ (see, for instance, \cite[p.435]{Meyer00}).

    \begin{figure}[t]
      \centering
      \includegraphics[width=0.5\textwidth]{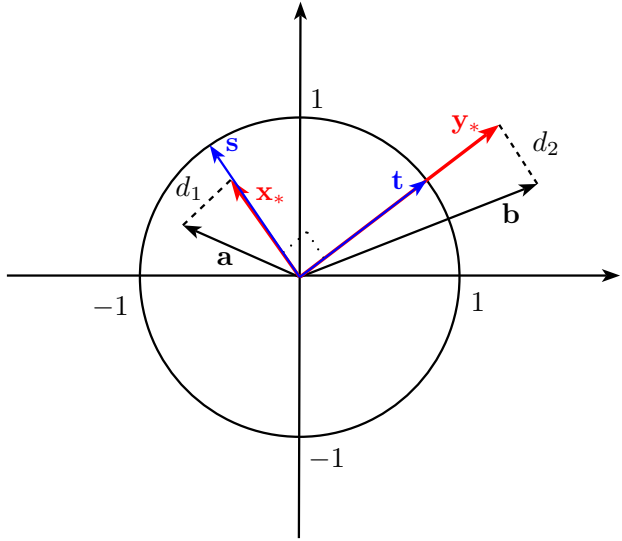}
      \caption{\scriptsize Geometric interpretation of the Pl\"{ucker} correction problem. Given vectors ${\mathbf a}$ and ${\mathbf b}$, the goal is to find two perpendicular vectors ${\mathbf x}_\ast$ and ${\mathbf y}_\ast$ that minimize the sum $d_1+d_2$ representing the sum of the distance between ${\mathbf a}$ and the line through the origin with the direction of the unit vector ${\mathbf s}$ with the distance between ${\mathbf b}$ and the line through the origin with the direction of the unit vector ${\mathbf t}$.}
      \label{figure1}
    \end{figure}

    Hence, the constrained problem (\ref{min-prob}) corresponds to find orthonormal vectors ${\mathbf s}$ and ${\mathbf t}$ minimizing the sum of the distance $d_1$, between ${\mathbf a}$ and the line through the origin with the direction of ${\mathbf s}$, with the distance $d_2$, between ${\mathbf b}$ and the line through the origin with the direction of ${\mathbf t}$ (see Fig.~\ref{figure1}). This means that (\ref{min-prob}) can be reformulated as
    \begin{equation}\label{min-prob2}
      \min_{\begin{array}{c}
        {\mathbf s}^T{\mathbf t}={\mathbf 0}, \\
        \|{\mathbf s}\|=1,\ \|{\mathbf t}\|=1
      \end{array}
    } \left\|[{\mathbf a}\ {\mathbf b}]-\left[({\mathbf a}^T{\mathbf s}){\mathbf s}\quad ({\mathbf b}^T{\mathbf t}){\mathbf t}\right]\right\|^2.
  \end{equation}
  This formulation of the Pl\"{ucker} correction problem is apparently more complicated and less practical than (\ref{min-prob}), but it is helpful to show that (\ref{min-prob}) has in fact a global minimum. To see this, one just needs to observe that the constraints in (\ref{min-prob2}) define a closed and bounded (and, consequently, compact) set in $\mathbb{R}^{2n}$ endowed with the Euclidean metric. Thus, by the classical Weierstrass theorem (see, for instance, \cite[Appendix A.6]{Luenberger08}), there exists at least a global minimum. This proves that the analysis carried out in Section \ref{local}, using the Lagrange multipliers, guarantees that the objective function (\ref{obj-function}) attains a global minimum at
  $$\left({\mathbf x}_2,{\mathbf y}_2\right)=\left(\frac{{\mathbf a}-\lambda_2{\mathbf b}}{1-\lambda_2^2},\, \frac{{\mathbf b}-\lambda_2{\mathbf a}}{1-\lambda_2^2}\right),$$
  where $\lambda_2$ is defined as in (\ref{lambda}).

  \subsection{Cases ${\mathbf a}= \pm {\mathbf b}$}

  As far as we know, the cases when ${\mathbf a}= \pm {\mathbf b}$ rarely occur in practical problems of computer vision. However, it is worth to make some comments on this particular case.

  If ${\mathbf a}={\mathbf b}$ with ${\mathbf b}={\mathbf 0}$, the solution of (\ref{min-prob}) is obviously ${\mathbf x}={\mathbf 0}$ and ${\mathbf y}={\mathbf 0}$. If ${\mathbf a}={\mathbf b}$ but ${\mathbf b}\neq {\mathbf 0}$ then replacing ${\mathbf b}$ by ${\mathbf a}$ in the Lagrangian (\ref{lagrangian}), we get a simpler expression. Finding the first order optimality conditions and solving them, one easily concludes that any pair of vectors of the form $({\mathbf a}+{\mathbf y},{\mathbf y})$, with $({\mathbf a}+{\mathbf y})^T{\mathbf y}={\mathbf 0}$ and ${\mathbf y}$ arbitrary, gives a local minimum. The value of the Lagrangian at all these local minima is $\|{\mathbf a}\|^2$ and does not depend on ${\mathbf y}$. Choosing, for instance, ${\mathbf y}={\mathbf 0}$, it follows that the pair $({\mathbf a},{\mathbf 0})$ is a local minimum of (\ref{min-prob}). Using a similar argument to that of Sec. \ref{global}, this pair is also a global minimum.

  Similarly, if ${\mathbf a}=-{\mathbf b}$, with ${\mathbf b}\neq {\mathbf 0}$, it can be concluded that $({\mathbf a},{\mathbf 0})$ provides also a global minimum for (\ref{min-prob}).

  \section{Experiments}\label{experiments}

%   \begin{figure}[t]
%     \centering
%     \includegraphics[width=0.48\textwidth]{figure2a.eps}
%     \caption{\scriptsize This Figure shows the evaluation of the method proposed in this paper {\tt LMPC} against state-of-the-art method {\tt BS}, proposed at~\cite{Bartoli05}, using $10^8$ randomly generated trials.}
%     \label{figure2}
%   \end{figure}

  In this section, we compare the method based on the explicit formula (\ref{eq:explicit:solution}), with the method of Bartoli and Sturm~\cite{Bartoli05}, in terms of 
  %\deleted{numerical results and}
   computational effort. Both methods were implemented using {\tt MATLAB}. The codes will be available in the author's website.
  We consider three different algorithms:
  \begin{itemize}
    \item {\tt LMPC} which denotes to the method derived in this paper;
    \item{\tt BS} which corresponds to the Bartoli and Sturm's approach; and
    \item{\tt BS-LSVD} which denotes to the method proposed by Bartoli and Sturm, where the SVD is computed using closed-form techniques.
  \end{itemize}
  A detailed description of each algorithm is shown in~\ref{app:alagorithms}

  %\deleted{Two experiments were carried out: the distribution of the error on the Pl\"{u}cker correction for general unconstrained 6-dimensional vectors, and evaluation of the error as a function of the deviation from the orthogonal constraint. The performance of the methods was investigated by means of numerical errors and computational effort. {\tt LMPC} denotes the ``Lagrange Multipliers Pl\"{u}cker Correction'' method proposed in this paper; and {\tt BS} stands for the method of Bartoli and Sturm~\cite{Bartoli05}.}

  To compare the methods we use the following procedure: we randomly generated unconstrained $10^{6}$ vectors $\mathbf{a},\,\mathbf{b}\in\mathbb{R}^3$ (Klein quadric $\mathbf{a}^T \mathbf{b} = 0$ is not enforced). For each trial, we apply both Pl\"{u}cker correction algorithms to the respective six-dimensional vectors, storing the values of the corresponding objective functions (\ref{obj-function}). 
  %\deleted{Fig.~\ref{figure2} displays the number of occurrences in terms of the values of the objective functions. This experiment was implemented in C++. The computational time for all the $10^6$ trials using our method was $16.6$ [s], while for the method of Bartoli \& Sturm was $27.3$[s].}
   The results in terms of computational time required for each algorithm is shown in Tab.~\ref{tab:experimental_results}.
  \begin{table}
    \caption{Evaluation of the computational time for the three algorithms, for general unconstrained $10^6$ trials.}\label{tab:experimental_results}
    \centering
    \begin{tabular}{|c|c|c|}
      \hline
      {\bf Algorithm} & For all trials & For each trial (median) \\ \hline \hline
      {\tt LMPC} & 6.5797 $s$ & 5.3940 $\mu s$ \\ \hline
      {\tt BS} & 43.688 $s$ & 34.827  $\mu s$ \\ \hline
      {\tt BS-LSVD} & 14.0058 $s$ & 11.805 $\mu s$ \\ \hline
    \end{tabular}
  \end{table}

 Both {\tt LMPC} and {\tt BS-LSVD} methods can be implemented using only closed-form steps. However, while {\tt LMPC} can be computed with a few steps (it only requires six lines of code), {\tt BS-LSVD} requires more algebraic operations and takes a longer time to run, see Tab.~\ref{tab:experimental_results}. Indeed, {\tt BS-LSVD} is about two times slower than {LMPC}. On the other hand, if we consider the original Bartoli and Sturm approach {\tt BS}, which requires the computation of two singular value decompositions, it requires iterative steps and, as a result, this method is significantly slower. From our experiments, one can see that this method is more than six times slower than our approach.

 \subsection{Discussion of the Experimental Results}

  In many applications there are estimates of 3D line coordinates that are not obtained from points. 3D lines can be estimated from intersections of planes, for example, or they can be obtained from non-conventional sensors, such as non-central generic cameras. In non-central generic cameras calibration, it consists in estimating a 3D line for each pixel~\cite{article:sturm:2011,article:miraldo:2013}. If we represent 3D lines using Pl\"{u}cker coordinates, the Klein quadric constraint has to be enforced --- in this case this correction has to be applied to all and each pixel, which corresponds to call the Pl\"{u}cker correction algorithm a large number of times. Let us consider a camera system, with a standard image size of $1280\times 1024$, which contains a total of $1310720$ pixels. For this case and from the experimental results, one can conclude that the Pl\"{u}cker correction step using the method proposed in this paper will be, at least, two times faster than Bartoli and Sturm method, which consists in saving more than eight seconds.

Other example is, using perspective cameras, the estimation of 3D lines from the intersections of the back-projecting planes from two or more images.

  \section{Conclusions}\label{sec:conclusions}

  In this paper we have addressed the Pl\"{u}cker correction problem, by minimizing the Frobenius norm between the estimated and input vectors. By solving the corresponding optimization problem using Lagrange multipliers, a simple solution, that can be computed in closed-form with a very small number of operations, was proposed. 
 % \deleted{While the state-of-the-art method does not compute a global minimum in general}
  Contrarily to the state-of-the-art method, we have proved theoretically that our method computes always the global minimum. In addition, the special cases (where a solution cannot be computed) are analysed. As the experimental results show, the proposed method is faster. 
   %\deleted{and gives lower values for the objective function}.

  \appendix
  \section{Algorithms}
  \label{app:alagorithms}
  In this section we show the code used in the experiments. Considering two general vectors $\mathbf{a} = \left(a_1, a_2, a_3\right)$ and $\mathbf{b} = \left(b_1, b_2, b_3\right)$ that do not verify the {\it Klein constraint}, using our method, the closest orthogonal vectors are given by $\mathbf{x} = \left(x_1, x_2, x_3\right)$ and $\mathbf{y} = \left(y_1, y_2, y_3\right)$ such that:

  \begin{Verbatim}[gobble=2,numbers=left,frame=lines,label=LMPC]
    p = a1*b1 + a2*b2 + a3*b3;
    q = a1*a1 + a2*a2 + a3*a3 + b1*b1 + b2*b2 + b3*b3;
    mu = 2*p/(q+sqrt(q*q-4*p*p));
    u_ = 1/(1-mu*mu);
    x1 = (a1-mu*b1)*u_; x2 = (a2-mu*b2)*u_; x3 = (a3-mu*b3)*u_;
    y1 = (b1-mu*a1)*u_; y2 = (b2-mu*a2)*u_; y3 = (b3-mu*a3)*u_;
  \end{Verbatim}

    The algorithm proposed by Bartoli and Sturm is based on the singular value decomposition and can be implemented as:
  \begin{Verbatim}[gobble=2,numbers=left,frame=lines,label=BS]
    [U,S,V] = svd(A,0);
    Z = S*V';
    z11 = Z(1,1); z21 = Z(2,1); z22 = Z(2,2); z12 = Z(1,2);
    T = [z12, z22; z21, -z11];
    [~,St,V_] = svd(T);
    hv = V_(:,2);
    hV = [hv(1),-hv(2);hv(2),hv(1)];
    R = U*hV*diag(diag(hV'*S*V'));
    x = R(:,1);
    y = R(:,2);
  \end{Verbatim}

  Usually, the singular value decomposition requires iterative techniques. However, since in this case we are dealing with $3\times 3$ matrices, is it possible to derive a closed form solution for this decompositions. Thus, in our experiments we also implemented  a closed form solution for the Bartoli and Sturm method:
  \begin{Verbatim}[gobble=2,numbers=left,frame=lines,label=BS-LSVD]
    s1 = sqrt((a11*a11*a11*a11 + 2*a11*a11 *a12*a12 + 2*a11*a11 *a21*a21 - . . . )
    s2 = sqrt(a11*a11 /2 - (a11*a11*a11*a11 + 2*a11*a11 *a12*a12 + . . . );
    s2 = sqrt(s2);
    v11 = -((a11*a12 + a21*a22 + a31*a32))/(a11^2 + a21^2 + a31^2 - s1*s1);
    v21 = 1;
    nv = (v11*v11 + v21*v21)^(1/2);
    v11 = v11/nv; v21 = v21/nv;
    v12 = v21; v22 = -v11;
    u11 = (a12*v21 + a11*v11)/s1; u12 = (a12*v22 + a11*v12)/s2;
    u21 = (a21*v11 + a22*v21)/s1; u22 = (a21*v12 + a22*v22)/s2;
    u31 = (a31*v11 + a32*v21)/s1; u32 = (a31*v12 + a32*v22)/s2;
    z11 = s1*v11; z12 = s1*v21;
    z21 = s2*v12; z22 = s2*v22;
    % t11 = z21; t12 = z22; t21 = z12; t22 = -z11;
    t11 = z12; t12 = z22; t21 = z21; t22 = -z11;
    st1 = t11/2 + t22/2 - (t11*t11 - 2*t11*t22 + t22*t22 + 4*t12*t21)^(1/2)/2;
    st2 = t11/2 + t22/2 + (t11*t11 - 2*t11*t22 + t22*t22 + 4*t12*t21)^(1/2)/2;
    if st1 < st2
        v1 = (t12*1)/(st1 - t11); v2 = 1; nv = (v1*v1 + v2*v2)^(1/2);
        v1 = v1/nv; v2 = v2/nv;
    else
        v1 = (t12*1)/(st2 - t11); v2 = 1; nv = (v1*v1 + v2*v2)^(1/2);
        v1 = v1/nv; v2 = v2/nv;
    end
    h11 = v1; h12 = -v2; h21 = v2; h22 = v1;
    x1 = (u11*h11 + u12*h21)*(h11*s1*v11 + h21*s2*v12);
    y1 = (u11*h12 + u12*h22)*(h12*s1*v21 + h22*s2*v22);
    x2 = (u21*h11 + u22*h21)*(h11*s1*v11 + h21*s2*v12);
    y2 = (u21*h12 + u22*h22)*(h12*s1*v21 + h22*s2*v22);
    x3 = (u31*h11 + u32*h21)*(h11*s1*v11 + h21*s2*v12);
    y3 = (u31*h12 + u32*h22)*(h12*s1*v21 + h22*s2*v22);
  \end{Verbatim}

  As it can be easily seen in the previous codes, our method {\tt LMPC} requires much less operations when compared with the closed form of Bartoli and Sturm algorithm {\tt BS-LSVD}. On the other hand, it does not require iterative techniques, such as the {\tt BS} method, which usually makes the algorithm slower.

  \end{document}